\documentclass[letterpaper, 10 pt, conference]{ieeeconf}  

\IEEEoverridecommandlockouts                              

\usepackage[
    style=ieee,
    doi=false,
    isbn=false,
    url=false,
    eprint=false,
    backend=biber,
    natbib=true
    ]{biblatex}
\usepackage[american]{babel}
\usepackage{csquotes}
\usepackage{graphicx}

\usepackage{pdf14}

\renewcommand{\cite}{\parencite}
\usepackage[keeplastbox]{flushend}
\usepackage{booktabs}
\let\labelindent\relax 
\usepackage[inline]{enumitem}

\usepackage[font=small]{subcaption}
\usepackage[font=small]{caption}
\usepackage{amsmath}
\usepackage{amsfonts}
\usepackage{csquotes}
\usepackage{perl_acronyms}
\usepackage{perl_misc}

\usepackage{calc}

\title{\LARGE \bf
Driving in the Matrix: Can Virtual Worlds Replace Human-Generated Annotations for Real World Tasks?
}

\author{Matthew Johnson-Roberson$^{1}$, Charles Barto$^{2}$, Rounak Mehta$^{3}$, Sharath Nittur Sridhar$^{2}$,\\ Karl Rosaen$^{2}$, and Ram Vasudevan$^{4}$
\thanks{$^{1}$M. Johnson-Roberson is with the Department of Naval Architecture and Marine Engineering, University of Michigan, Ann Arbor, MI 48109 USA {\tt\small mattjr@umich.edu}}%
\thanks{$^{2}$ C. Barto, S. Sridhar, and K. Rosaen are with the Electrical Engineering and Computer Science, University of Michigan, Ann Arbor, MI 48109 USA {\tt\small bartoc,sharatns,krosaen@umich.edu}}
\thanks{$^{3}$ R. Mehta is with the Robotics Program, University of Michigan, Ann Arbor, MI 48109 USA {\tt\small rounak@umich.edu}}
\thanks{$^{4}$ R. Vasudevan is with the Mechanical Engineering, University of Michigan, Ann Arbor, MI 48109 USA {\tt\small ramv@umich.edu}}
}
\begin{document}

\maketitle
\thispagestyle{empty}
\pagestyle{empty}

\begin{abstract}
Deep learning has rapidly transformed the state of the art algorithms used to address a variety of problems in computer vision and robotics.
These breakthroughs have relied upon massive amounts of human annotated training data. 
This time consuming process has begun impeding the progress of these deep learning efforts. 
This paper describes a method to incorporate photo-realistic computer images from a simulation engine to rapidly generate annotated data that can be used for the training of machine learning algorithms.
We demonstrate that a state of the art architecture, which is trained \emph{only} using these synthetic annotations, performs better than the identical architecture trained on human annotated real-world data, when tested on the KITTI data set for vehicle detection. 
By training machine learning algorithms on a rich virtual world, real objects in real scenes can be learned and classified using synthetic data. This approach offers the possibility of accelerating deep learning's application to sensor-based classification problems like those that appear in self-driving cars. 
The source code and data to train and validate the networks described in this paper are made available for researchers.
\end{abstract}


\section{Introduction}

The increasingly pervasive application of machine learning for robotics has led to a growing need for annotation. 
Data has proven to be both the limiting factor and the driver of advances in computer vision, particularly within semantic scene understanding and object detection~\cite{imagenet,vgg16}. A variety of approaches have been proposed to efficiently label large amounts of training data including crowdsourcing, gamification, semi-supervised labeling, and Mechanical Turk ~\cite{Russell:2008:LDW:1345995.1345999,Ahn:2004uq,Ahn:2006fk}. 
These approaches remain fundamentally bounded by the amount of human effort required for manual labeling or supervision.

To devise a strategy to address this ever-growing problem, this paper proposes the use of computer annotated photo-realistic simulations to train deep networks, enabling robots to semantically understand their surroundings.
In particular, we focus on the task of vehicle detection within images. 
Although this application is specialized, the insights developed here can cut across a variety of perception-related tasks in robotics.
This vehicle detection task also serves as a useful motivator since a human driver's ability to operate safely using only visual data illustrates there is sufficient signal in images alone to safely operate an automobile.
The goal of vision-only driving has become something of a Holy Grail in the autonomous vehicle community due to the low cost of sensors.
 
Currently this remains beyond the state-of-the-art of robotics, but several paradigms have been proposed. These lie on a spectrum from component-based systems, where combinations of codified rules and learned classifiers are applied to each of the individual sub-tasks involved in a complex action like driving, to end-to-end learning, in which sensor data is mapped directly to the action space of driving through learning from many examples. 
Deep-learning has become an extremely popular approach to solve many of the sub-problems in the componentized method~\cite{kitti,deepmpc-lenz-knepper-saxena-rss2015} and has even been proposed as the most plausible method to solve the end-to-end problem in self-driving cars if such a solution is even plausible~\cite{LeCun:2005aa}.

Despite the potential to address a variety of problems in the self-driving car space, deep learning, like many purely data-driven approaches, is prone to overfitting. 
In fact, given the large number of parameters in deep learning systems, they have proven particularly prone to data set bias~\cite{dataset-bias2011,Herranz:2016aa,Khosla:2012aa}.
While data sets have spurred impressive progress in the research community, the lack of generalizability remains an impediment to wide adoption and deployment on fielded robotic systems~\cite{dataset-bias2011}.
In fact, this paper illustrates that current successful approaches to the vehicle detection task are based on a handful of data sets and a training process that leads to poor cross data set performance, which restricts real-world utility.

On the other hand, one can ask whether the relatively finite (order of thousands of images) current hand-labeled data sets are sufficiently large and diverse enough to ever learn general models that enable vehicles to operate safely in the world?
To address this problem, we consider the same cross data set validation task, where a model is learned on a completely independent data set (i.e. different geographic regions, time of day, camera system, etc.) and tested on another. We propose and explore visual simulation as a solution to the issue of finite human labeled data sets. 

Computer graphics has greatly matured in the last 30 years and a robust market for entertainment, including CGI movies, video games, and virtual reality, has spurred great innovation in our ability to generate photo-realistic imagery at high speed. If networks trained with this type of synthetic annotated data can perform as well as networks trained with human annotated data on real-world classification tasks, then one can begin rapidly increasing the size of useful training data sets via simulation and verifying whether larger data sets are sufficient to learn general models. 

The contributions of this paper are as follows: \begin{enumerate*}
\item a fully automated system to extract training data from a sophisticated simulation engine;
\item experiments to address the issue of data set bias in deep learning-based object classification approaches, specifically in self-driving car data sets; 
\item state-of-the-art performance on real data using \textit{simulation only} training images; and
\item results that highlight improved performance with greater numbers of training images, suggesting the ceiling for training examples has not yet been reached for standard deep learning network architectures. 
\end{enumerate*}

The paper is organized as follows: \secref{s:background} presents prior work; \secref{s:system} describes the process used to collect the data and the training process; \secref{s:exp} discusses the experimental setup; \secref{s:res} presents preliminary results; \secref{s:disc} discusses these preliminary results in detail; and finally, \secref{s:con} presents our conclusions and future work.
\addtolength{\voffset}{0.2cm}

\section{Related Work}\label{s:background}

There have been several attempts to generate virtual data to boost the performance of deep learning networks. Early approaches used 3D renderings of objects to augment existing training data for pedestrian detection task~\cite{Marin:2010aa,Xu:2014aa}. Further work inserted synthetic pedestrians at a variety of angles into real scenes to train a part-based model classifier \cite{Hattori:2015aa}. \citet{Su:2015aa} employed 3D rendering to perform synthetic image generation for viewpoint estimation to construct training data for a \ac{CNN} based approach.

More recently, fully synthetic worlds have been suggested. Most relevant to this paper is the work of \citet{playing-for-data}, where the same engine we propose was used to generate simulated data for semantic segmentation of images. However, this work still required human annotators in the loop to supervise the generation of labels. Additionally, these data were not used alone but in concert with the real-world CamVid data set~\cite{camvid}.  A network trained on the combined dataset achieved superior performance to the same network trained on the CamVid training set alone, as evaluated on the CamVid testing dataset.

Each of these approaches refine the performance of their approach by fine tuning networks using a portion of the real-world testing data. This can be explained by a common problem in machine learning where networks trained on one data set suffer in terms of performance when evaluated on another data set \cite{dataset-bias2011}. This is of significant concern in the context of autonomous vehicles given that these algorithms need to run in different areas of the world and in a range of weather and lighting conditions.

An important exception to this training and evaluation approach is the recent semantic image-based segmentation result constructed using the SYNTHIA data set \cite{synthia-dataset}, which trained on 13,000 purely synthetic images. 
However, the authors relied upon a mixture of real and synthetic images to train a network that achieved comparable performance to a network trained only on real-world data.



\section{Technical Approach}\label{s:system}

This section describes our approach to generate synthetic images with computer-generated annotations of vehicles, which are then used to train our object detection network.

\subsection{Cloud-Cased Simulation Capture}\label{s:capture}
We endeavour to leverage the rich virtual worlds created for major video games to simulate the real world with a high level of fidelity. 
Specifically, we leverage Grand Theft Auto V (GTA V). The publisher of GTA V allows non-commercial use of footage from the game~\cite{playing-for-data}.
Data is captured from the game using two plugins, Script Hook V and Script Hook V.NET, developed by the open source community~\cite{scripthookv}. 
We refer to the first as the \textit{native plugin} and the second as the \textit{managed plugin}. 

\begin{figure*}[!t]
\centering
\begin{subfigure}{0.3\textwidth}
\includegraphics[width=\linewidth]{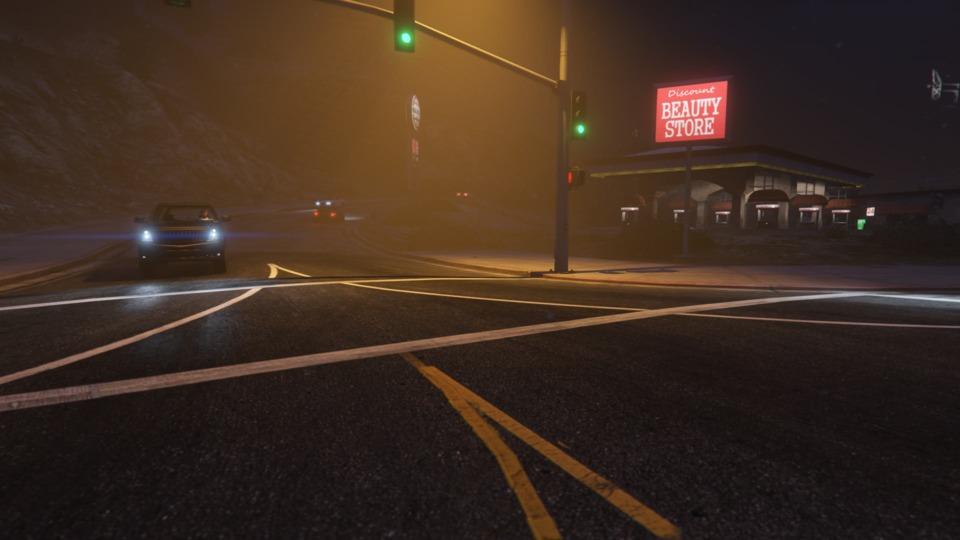}
\end{subfigure}
\hspace*{\fill}
\begin{subfigure}{0.3\textwidth}
\includegraphics[width=\linewidth]{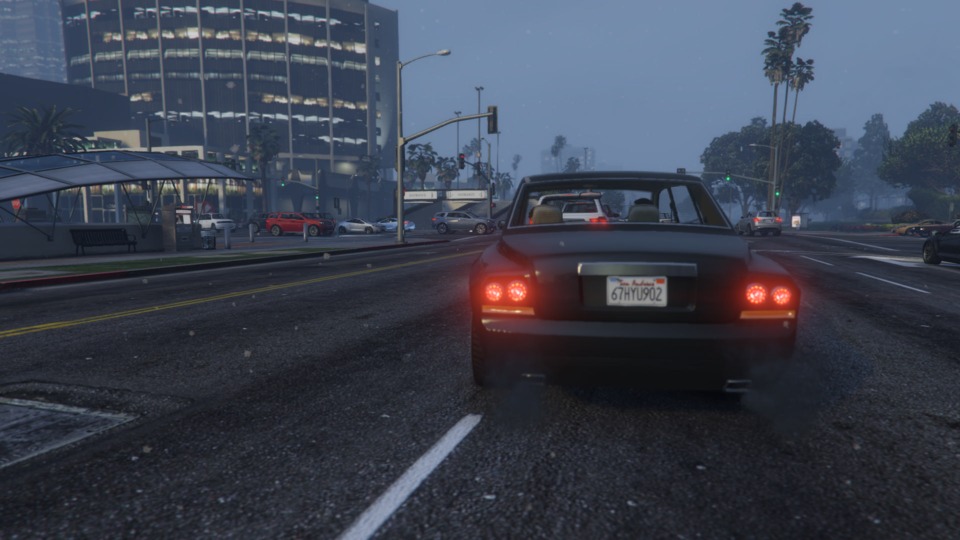}
\end{subfigure}
\hspace*{\fill} 
\begin{subfigure}{0.3\textwidth}
\includegraphics[width=\linewidth]{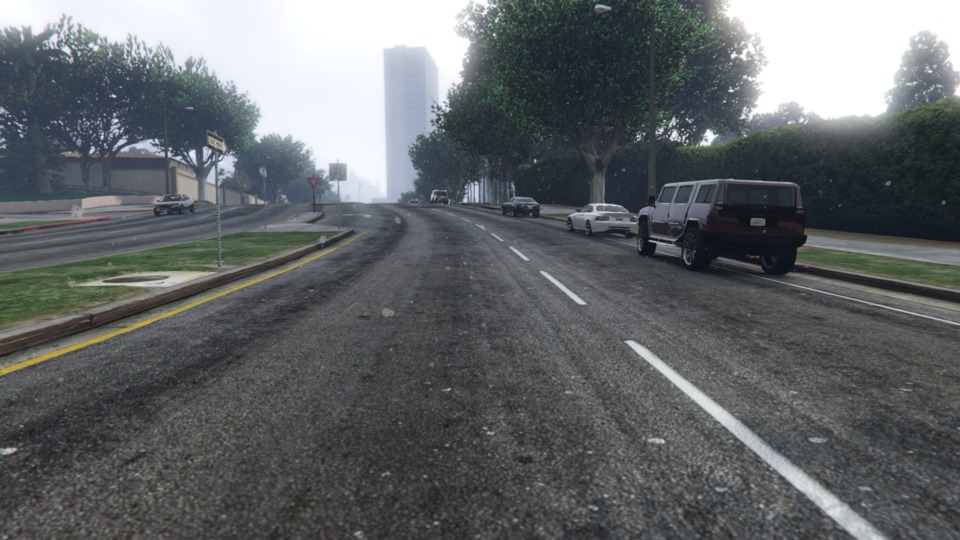}
\end{subfigure}

\bigskip
\begin{subfigure}{0.3\textwidth}
\includegraphics[width=\linewidth]{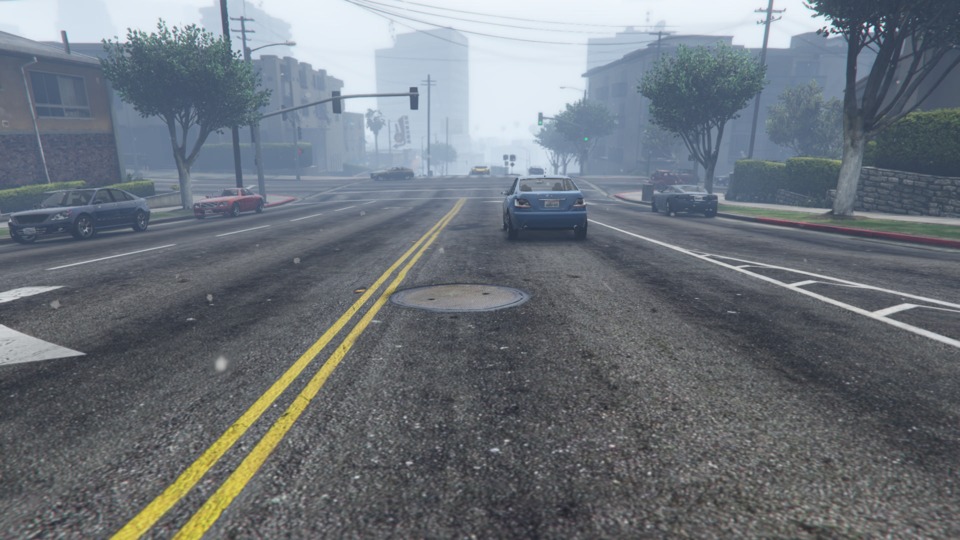}
\end{subfigure}
\hspace*{\fill} 
\begin{subfigure}{0.3\textwidth}
\includegraphics[width=\linewidth]{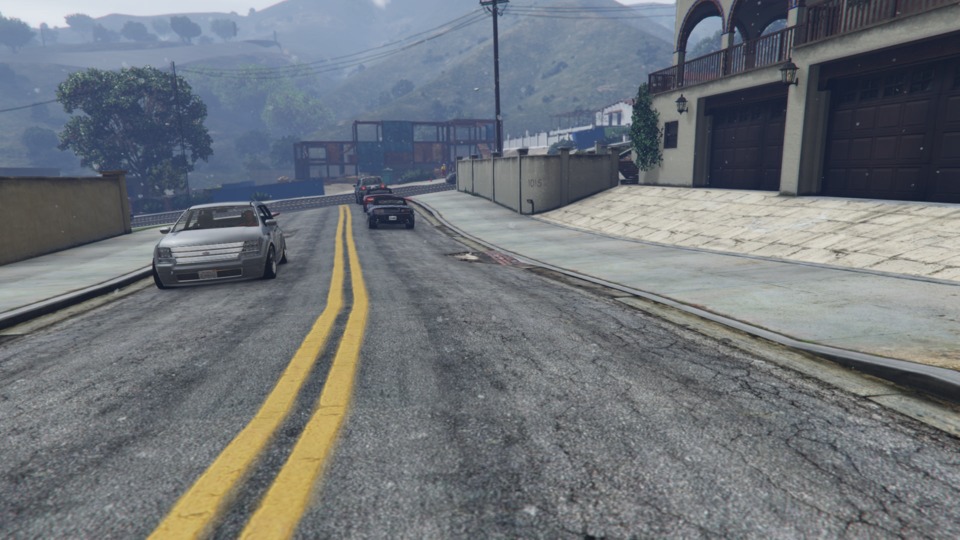}
\end{subfigure}
\hspace*{\fill} 
\begin{subfigure}{0.3\textwidth}
\includegraphics[width=\linewidth]{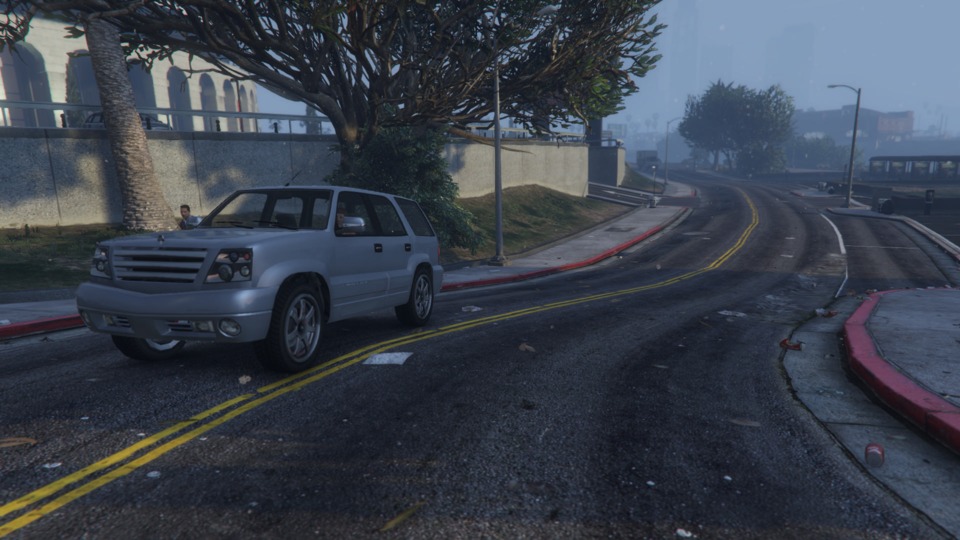}
\end{subfigure}

\bigskip
\begin{subfigure}{0.3\textwidth}
\includegraphics[width=\linewidth]{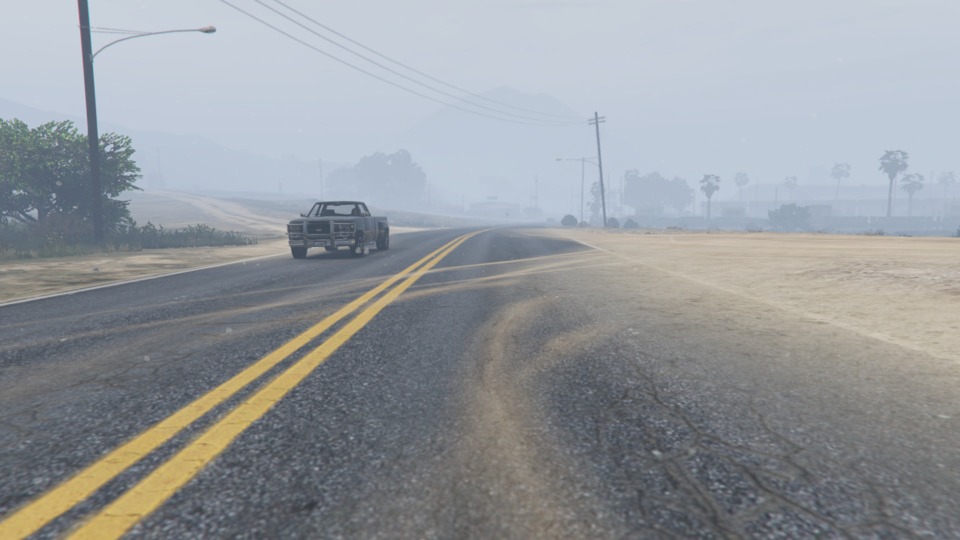}
\end{subfigure}
\hspace*{\fill} 
\begin{subfigure}{0.3\textwidth}
\includegraphics[width=\linewidth]{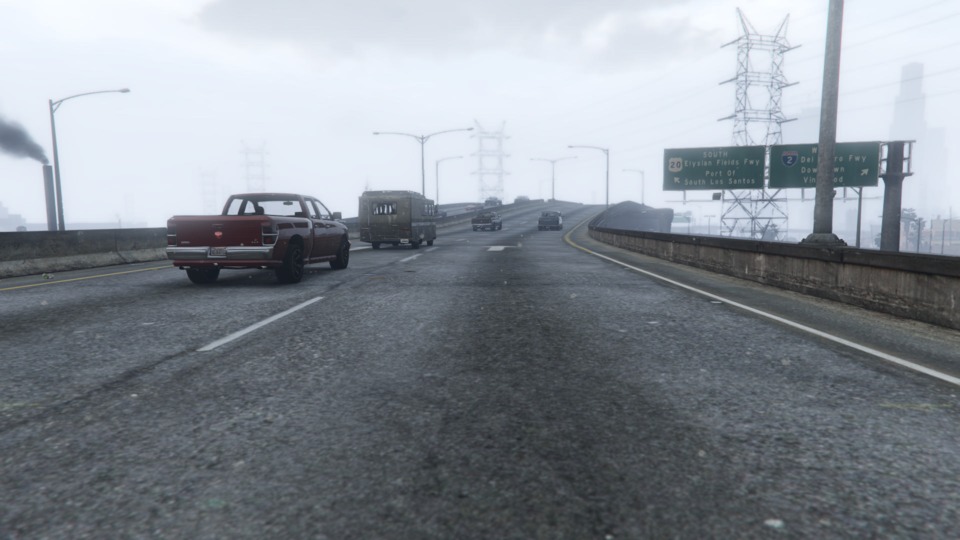}
\end{subfigure}
\hspace*{\fill} 
\begin{subfigure}{0.3\textwidth}
\includegraphics[width=\linewidth]{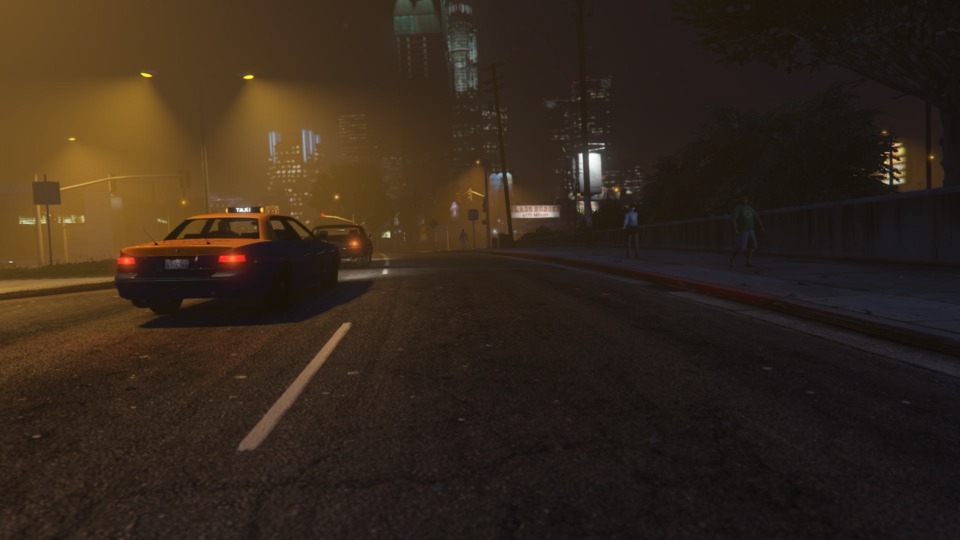}
\end{subfigure}

\bigskip
\hspace*{\fill} 

\hspace*{\fill} 
\caption{Sample images captured from the video game based simulation engine proposed in this paper. A range of different times of day are simulated including day, night, morning and dusk. Additionally the engine captures complex weather and lighting scenarios such as driving into the sun, fog, rain and haze.}
\label{f:gta}
\end{figure*}

\begin{figure*}[!t]
\centering
\begin{subfigure}{0.3\textwidth}
\includegraphics[width=\linewidth]{GTA_images/9769491xaaa.jpg}
\caption*{Clear}
\end{subfigure}
\hspace*{\fill} 
\begin{subfigure}{0.3\textwidth}
\includegraphics[width=\linewidth]{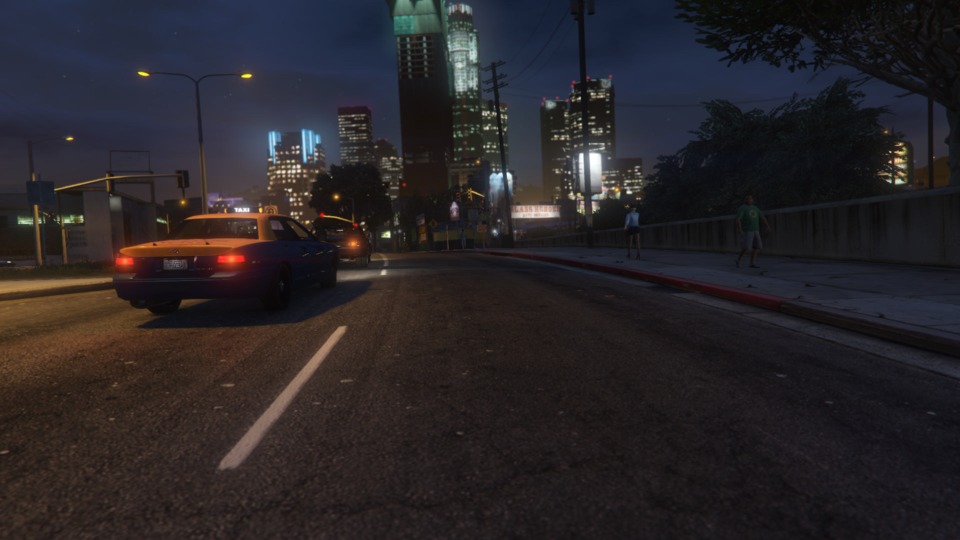}
\caption*{Overcast}
\end{subfigure}
\hspace*{\fill}
\begin{subfigure}{0.3\textwidth}
\includegraphics[width=\linewidth]{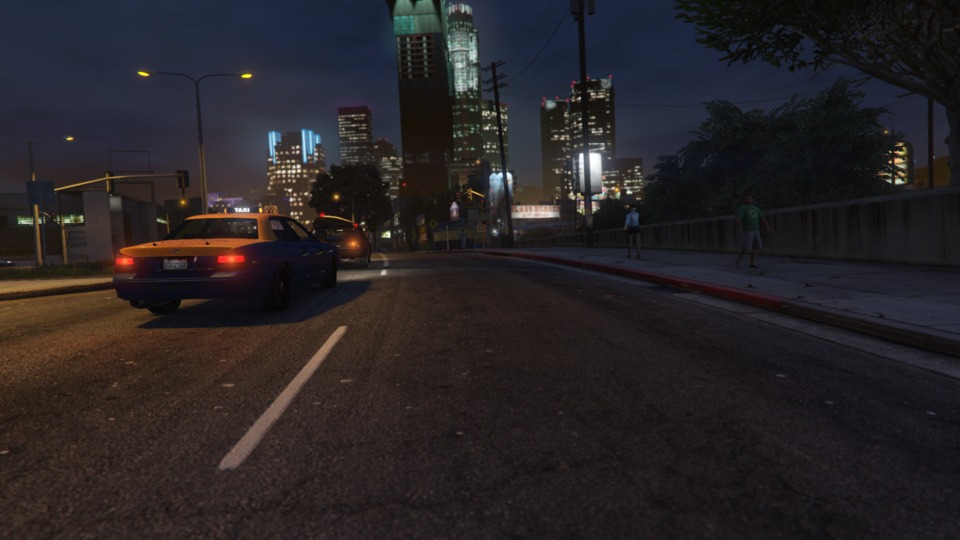}
\caption*{Raining}
\end{subfigure}
\bigskip
\begin{subfigure}{0.3\textwidth}
\includegraphics[width=\linewidth]{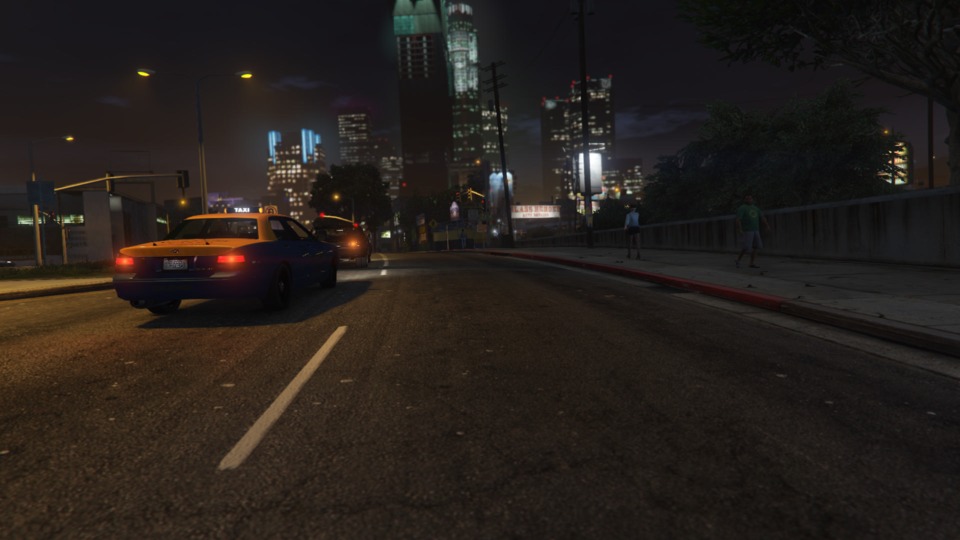}
\caption*{Light Snow}
\end{subfigure}
\hspace*{\fill}
\begin{subfigure}{0.3\textwidth}
\includegraphics[width=\linewidth]{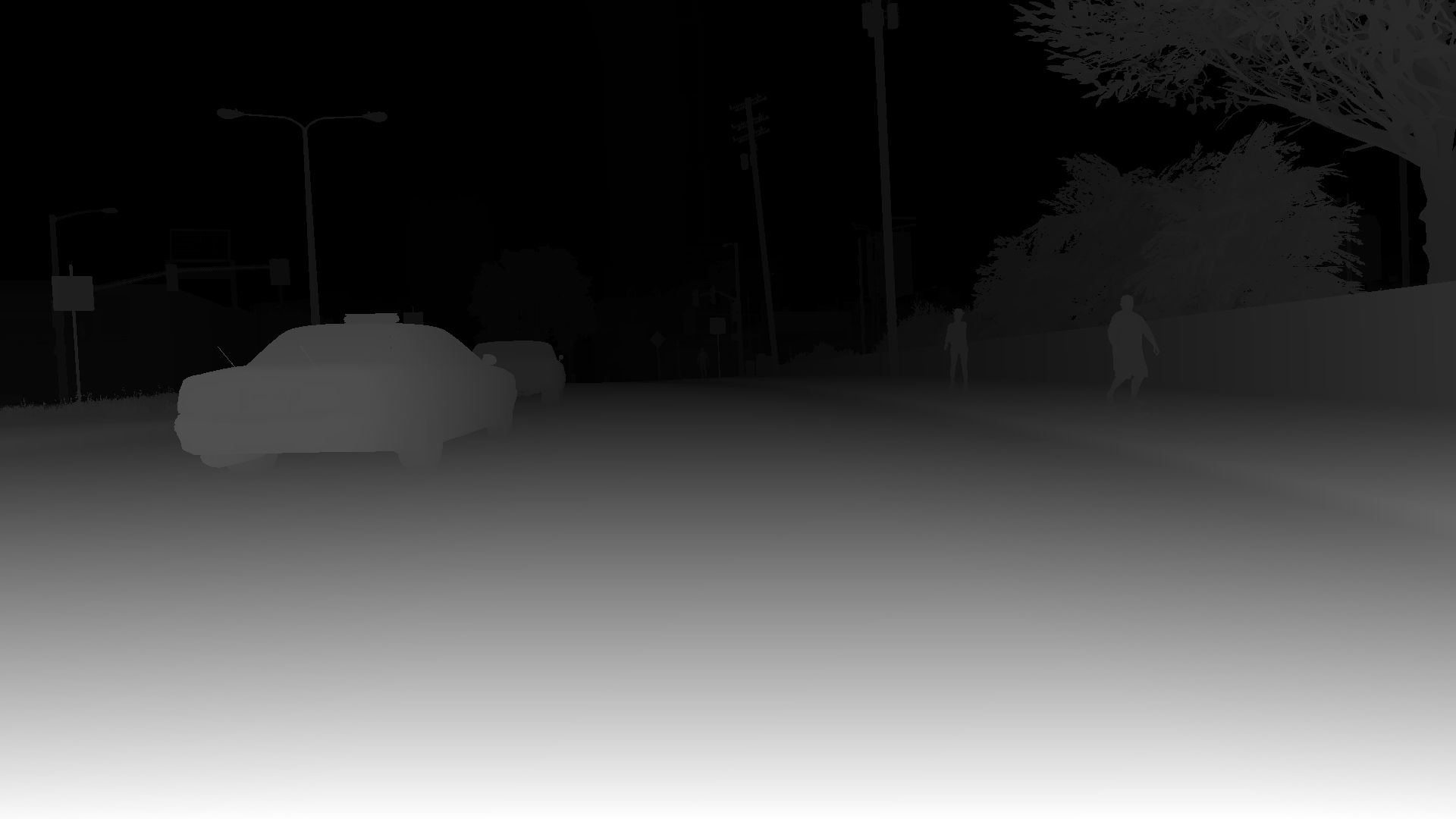}
\caption*{Scene Depth}
\end{subfigure}
\hspace*{\fill}
\begin{subfigure}{0.3\textwidth}
\includegraphics[width=\linewidth]{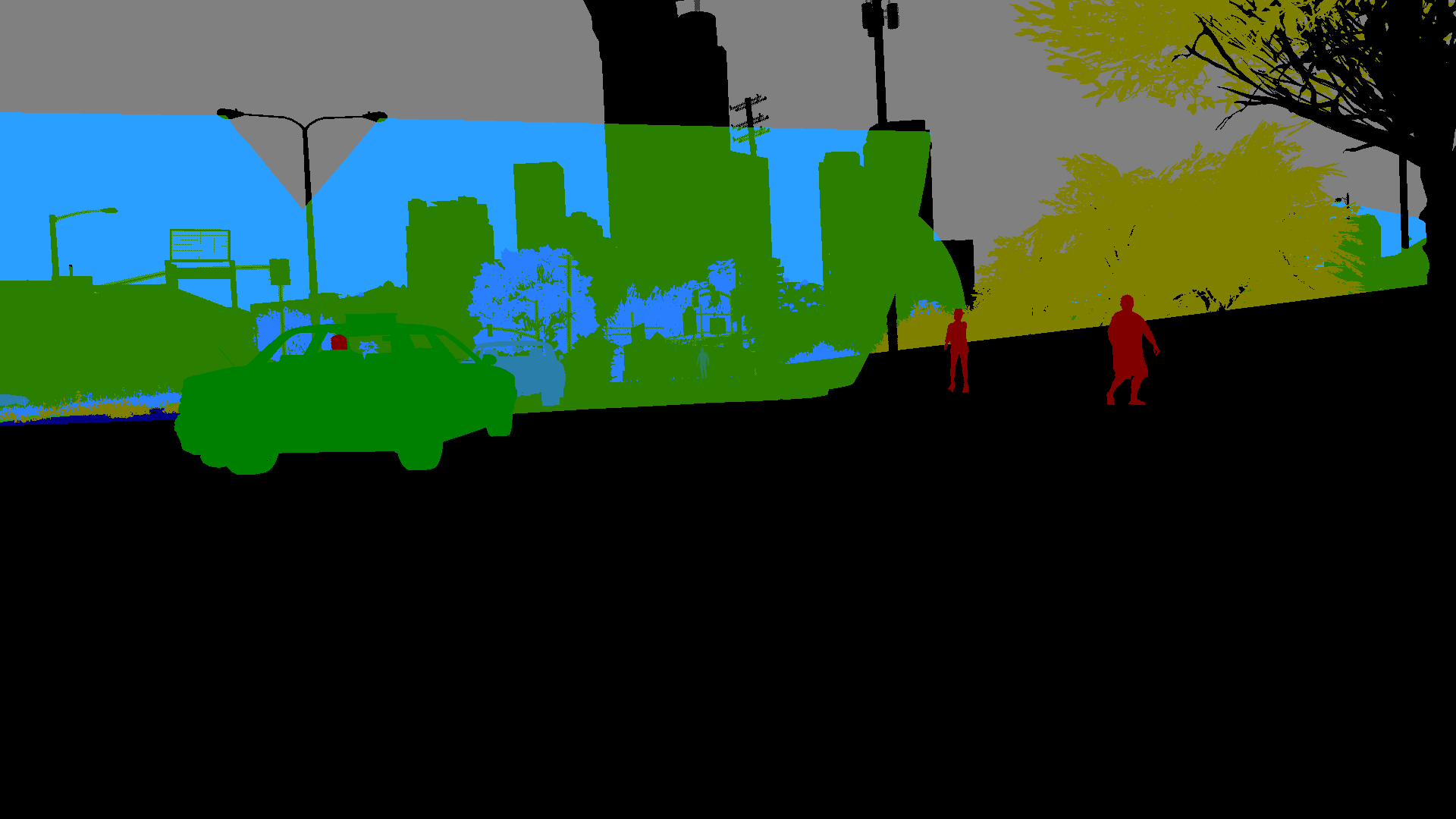}
\caption*{Pixel-wise Object Stencil Buffer}
\end{subfigure}
\caption{Four different weather types appear in the training data. The simulation can be paused and the weather condition can be varied. Additionally note the depth buffer and object stencil buffer used for annotation capture. In the depth image, the darker the intensity the farther the objects range from the camera. In the stencil buffer, we have artificially applied colors to the image's discrete values which correspond to different object labels for the game. Note that these values cannot be used directly. The process by which these are interpreted is highlighted in \protect\secref{s:buffers}. 
}
\label{f:weather}
\end{figure*}

Our process to generate simulated data is as follows: The managed plugin captures information about the scene at 1 Hz and uploads it to a cloud machine running an SQL server. In addition, it retrieves screen shots, scene depth, and auxiliary information from the game stored in the \ac{GPU}'s stencil buffer data. Sample images appear in \figref{f:gta}. As illustrated in \figref{f:weather}, for each simulation capture point we can save a maximum of five images using different weather types. 

Screen shots are captured by ``hooking" into Direct3D 11's present callback.
A process by which the native call is replaced with custom code which operates and then returns to the native call. 
In our custom code, we copy the graphics card's buffers. 
Specifically the depth and stencil data are captured by hooking into Direct3D's ID3D11ImmediateContext::ClearDepthStencilView and saving the buffers before each call. 
Because of optimizations applied by the graphics card drivers, we ``rehook" the clear function each frame. 
When saving each sample the managed plugin requests all current buffers from the native plugin and the buffers are downloaded from the \ac{GPU} and copied into managed memory. Buffers are saved in tiff format to zip archives and uploaded to a cloud server.

\subsection{Internal Engine Buffers}\label{s:buffers}

The format of the rendering engine's stencil and depth buffers are somewhat atypical. The value stored in the depth buffer is actually the logarithm of the actual depth. This enhances depth precision far away from the camera, and is one of the techniques that allows the engine to render such a large world. Since we are interested in depth ranges and not just depth ordering, we linearize the depth buffer before using it in post processing. 
The stencil buffer is an 8-bit per pixel buffer used to store information about each of the classes within a scene. 
It is utilized in the engine to store information about object class to answer queries such as ``which object is under the cross hair.'' 
The engine splits its stencil buffer in half, using the bottom four bits to store a numerical object ID, and the top four bits to store some flags. 

The managed plugin captures the 2D projection of each object's oriented bounding box, the position of each object, the position of the camera, and the object class. 
The projected bounding box, which we call the ground truth bounding box, is often loose so we post-process each ground truth detection using data from the stencil and depth buffers. 

\subsection{Tight Bounding Box Creation}\label{s:bbox}
\begin{figure*}
\centering
\begin{subfigure}{\textwidth}
\begin{subfigure}{0.3\textwidth}
\includegraphics[width=\linewidth]{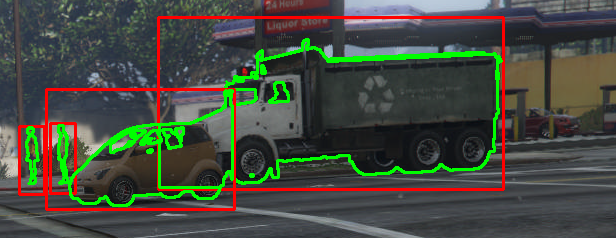}
\caption{}
\label{f:bbox-process:step1:color}
\end{subfigure}
\hspace*{\fill}
\begin{subfigure}{0.3\textwidth}
\includegraphics[width=\linewidth]{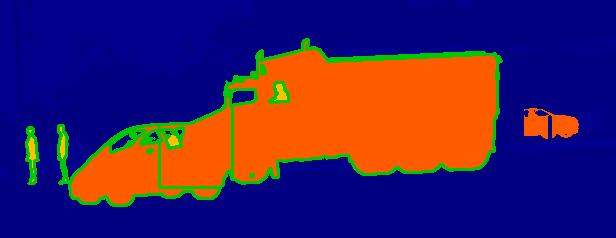}
\caption{}
\label{f:bbox-process:step1:stencil}
\end{subfigure}
\hspace*{\fill}
\begin{subfigure}{0.3\textwidth}
\includegraphics[width=\linewidth]{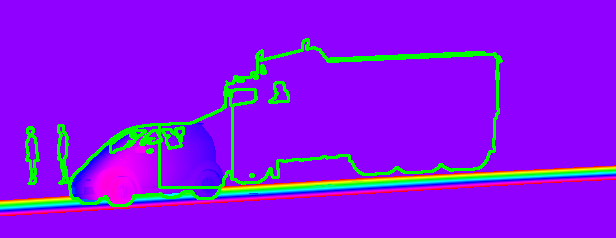}
\caption{}
\label{f:bbox-process:step1:depth}
\end{subfigure}
\label{f:bbox-process:step1}
\end{subfigure}
\bigskip
\begin{subfigure}{\textwidth}
\begin{subfigure}{0.3\textwidth}
\includegraphics[width=\linewidth]{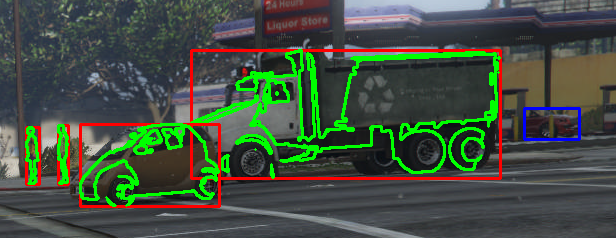}
\caption{}
\label{f:bbox-process:step2:color}
\end{subfigure}
\hspace*{\fill}
\begin{subfigure}{0.3\textwidth}
\includegraphics[width=\linewidth]{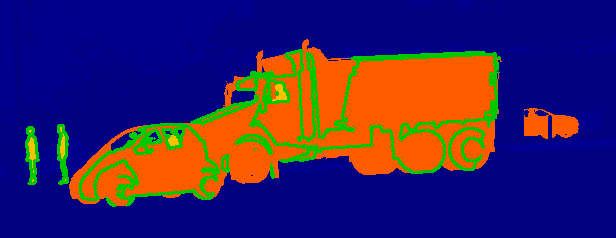}
\caption{}
\label{f:bbox-process:step2:stencil}
\end{subfigure}
\hspace*{\fill}
\begin{subfigure}{0.3\textwidth}
\includegraphics[width=\linewidth]{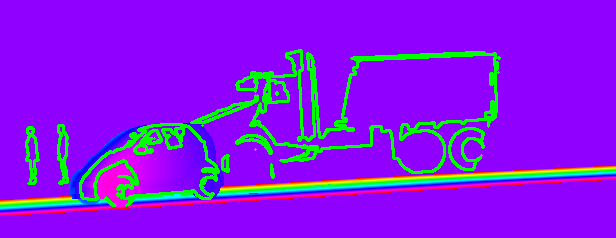}
\caption{}
\label{f:bbox-process:step2:depth}
\end{subfigure}
\label{f:bbox-process:step2}
\end{subfigure}
\caption{An illustration of the pipeline for tight bounding box creation. The engine's original bounding boxes are shown in (a). Since they are loose, we process them before using them as training data. In (b) we see an image from the stencil buffer, the orange pixels have been marked as vehicle enabling us to produce tight contours outlined in green. However, note that the two objects do not receive independent IDs in this pass so we must disambiguate the pixels from the truck and the compact car in a subsequent step. To do this we use the depth shown in (c) where lighter purple indicates closer range. This map is used to help separate the two vehicles, where (e) contains updated contours after processing using depth and (f) contains those same updated contours in the depth frame. Finally, (d) depicts the bounding boxes with the additional small vehicle detections in blue which are all used for training. Full details of the process appear in Section~\protect\ref{s:bbox}.}
\label{f:bbox-process}
\end{figure*}

The process to generate tight object bounding boxes without any human intervention is depicted at a high level in \figref{f:bbox-process}. The steps are as follows:
\begin{enumerate}
  \item The engine mantains coarse bounding boxes from the 3D positions of all objects in a fixed radius from the player. However these may be too loose to be directly useful for training (see Figure~\ref{f:bbox-process:step1:color}).
  \item To refine these boxes, contour detection is run on the stencil buffer (shown in green). As the stencil buffer contains simple pixel class labels, detections that partially occlude each other generate a single contour. This problem can be seen in \figref{f:bbox-process:step1:stencil}.
  \item Using the depth buffer we can compute the mean depth within the detected contour. See  \figref{f:bbox-process:step1:depth} and \figref{f:bbox-process:step2:depth} which shows depths around the mean for the compact car. 
  \item We then generate a new putative bounding box where included pixels are thresholded based upon a distance from the mean depth calculated in the previous step. The contours of accepted pixels are shown in Figs. \ref{f:bbox-process:step2:color} -- \ref{f:bbox-process:step2:depth}. Notice how the contours for the truck and the car are now distinct.
  We found that this improves the quality of the ground truth bounding boxes significantly. 
  \item Finally we also add bounding boxes of pixels classified as ``car'' in the stencil buffer that still do not have a corresponding ground truth bounding box. The blue bounding box in \figref{f:bbox-process:step2:color} is an example of one such detection. This allows us to include cars that are too far away to be registered with the physics engine but are still rendered.
\end{enumerate}

\subsection{Dataset Properties}
\begin{figure}[htb]
\centering
\begin{subfigure}{0.85\linewidth}
\includegraphics[width=\linewidth]{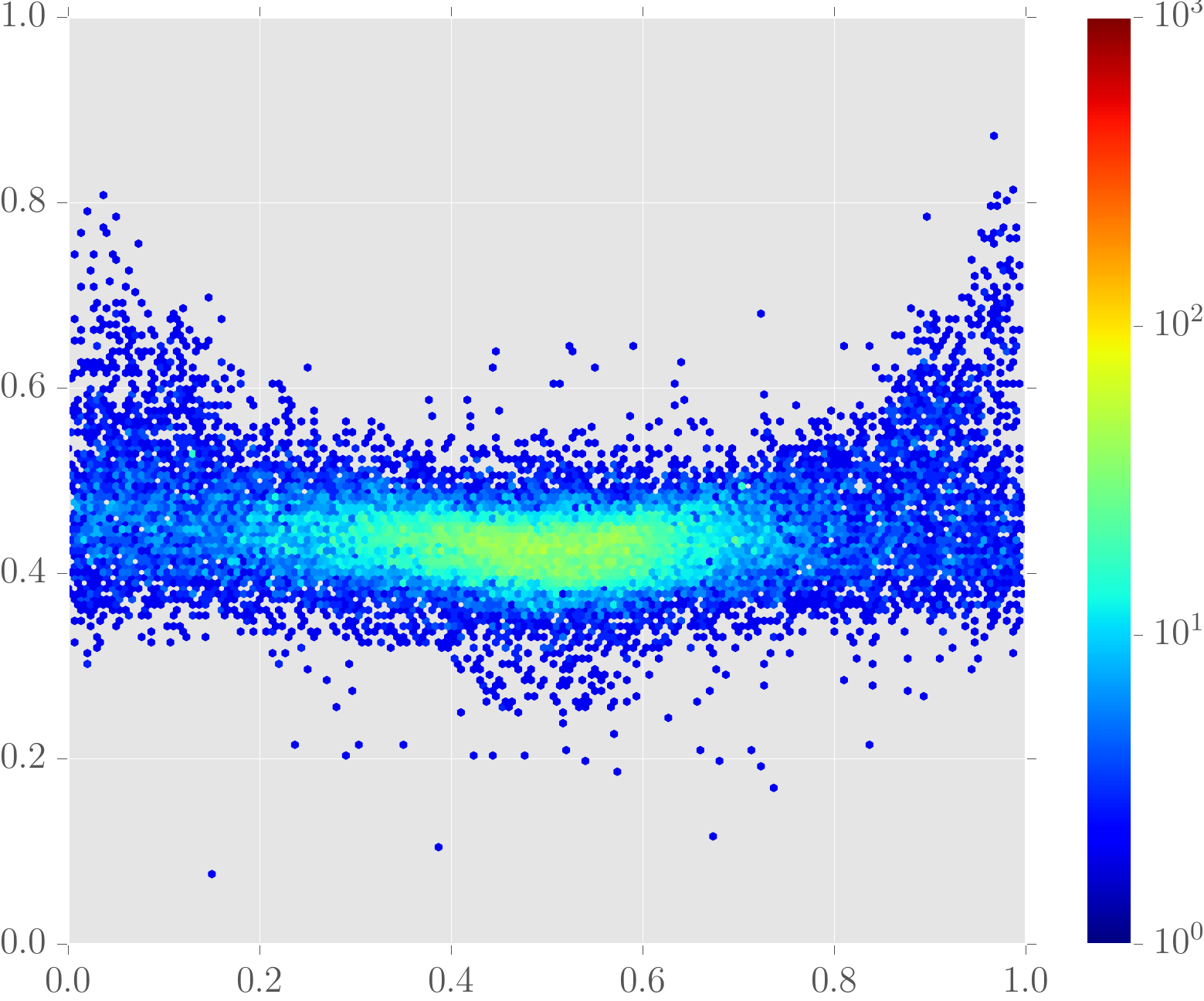}
\caption{Cityscapes}
\end{subfigure}
\bigskip
\begin{subfigure}{0.85\linewidth}
\includegraphics[width=\linewidth]{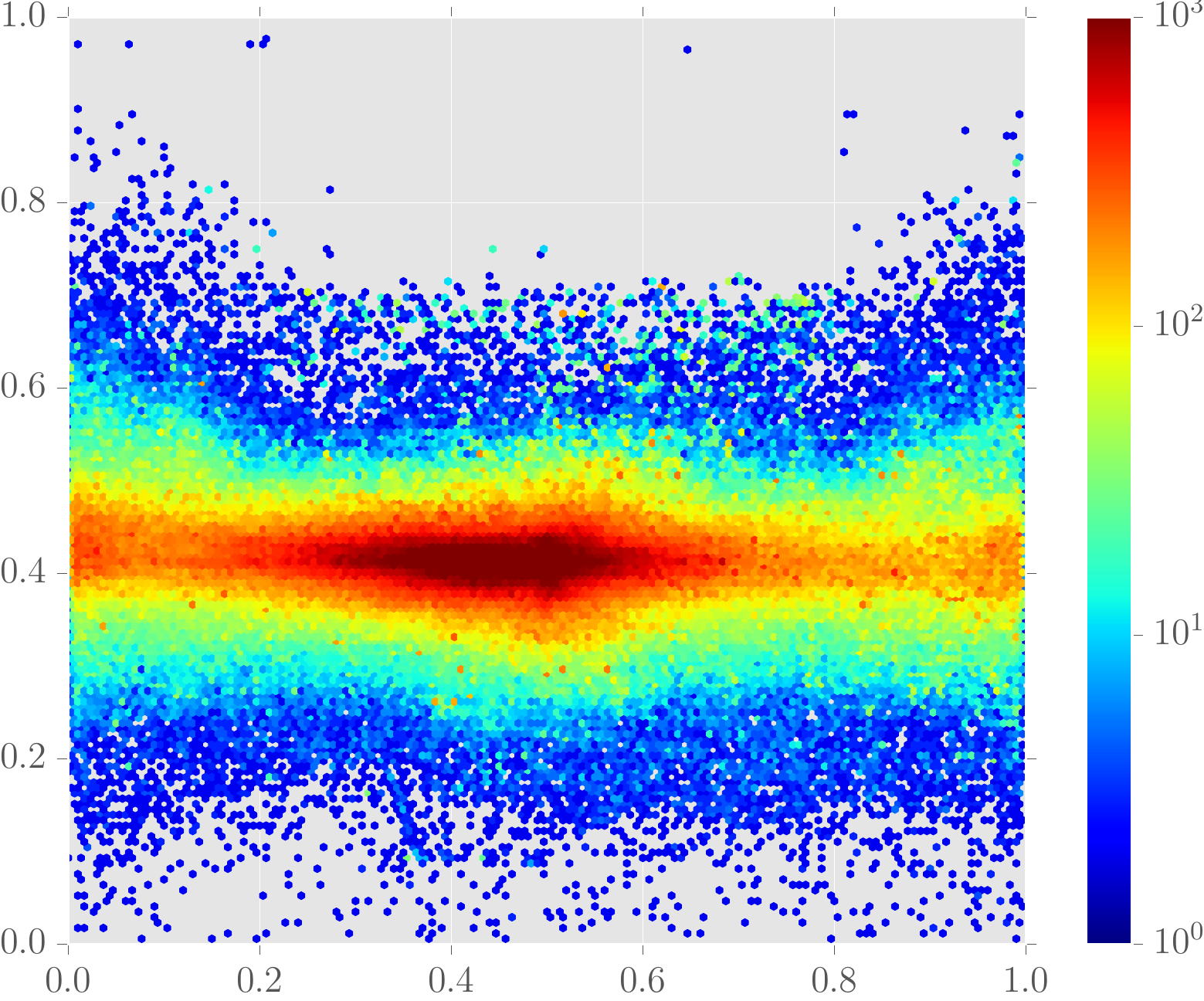}
\caption{Sim 200k}
\end{subfigure}
\caption{Heatmaps of the training data's bounding box centroids. These plots show the frequency of cars in different locations in the image. Note the much larger spread of occurrence location in the simulated data (b) than in the real images of Cityscapes (a). 
In the proposed approach, cars are found in a wide area across the image aiding the network in capturing the diversity of real appearance.}
\label{f:heatmap}
\end{figure}

\begin{figure*}[htb]
\centering
\begin{subfigure}{0.3\linewidth}
\includegraphics[width=\linewidth]{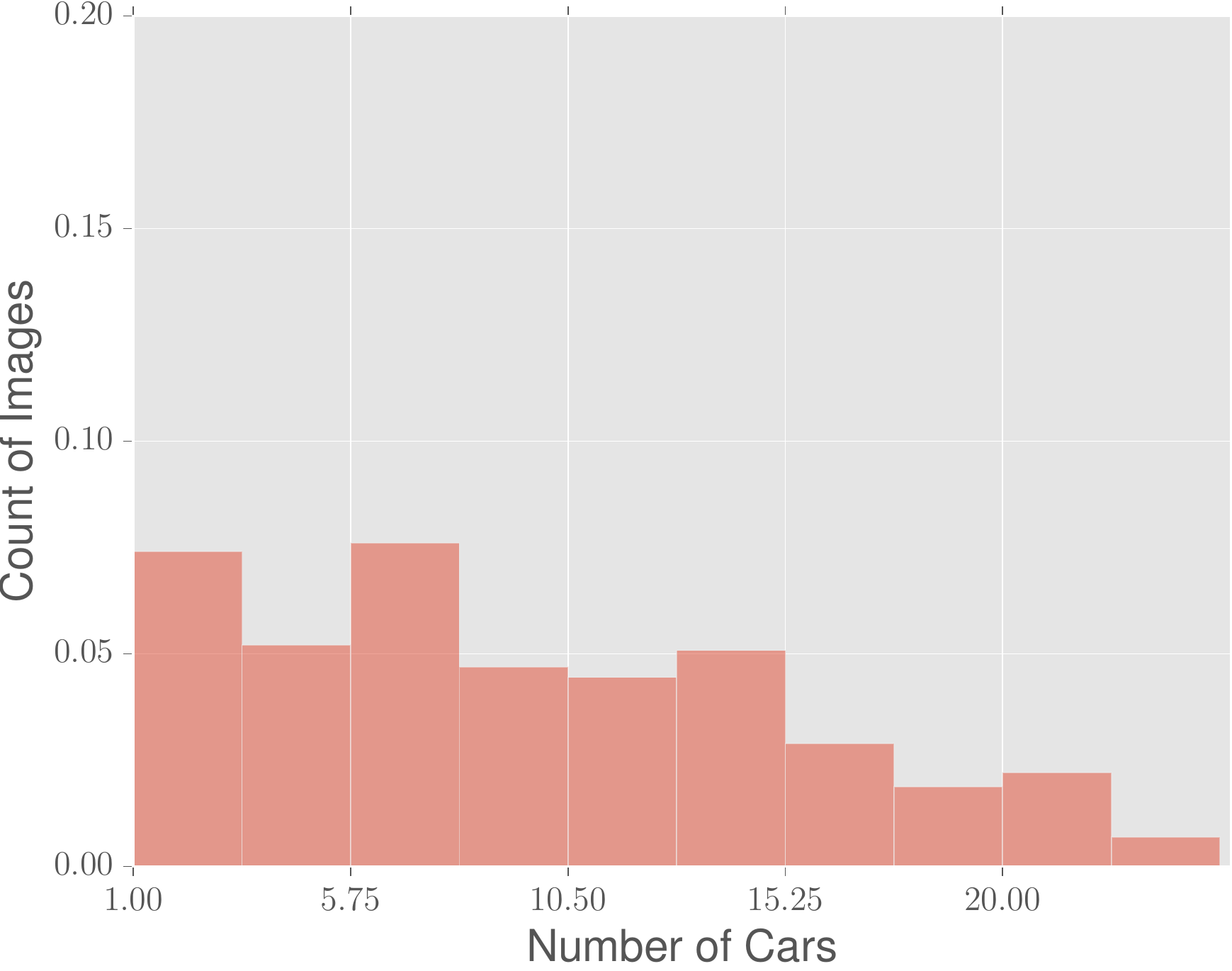}
\caption{Cityscapes}
\end{subfigure}
\begin{subfigure}{0.3\linewidth}
\includegraphics[width=\linewidth]{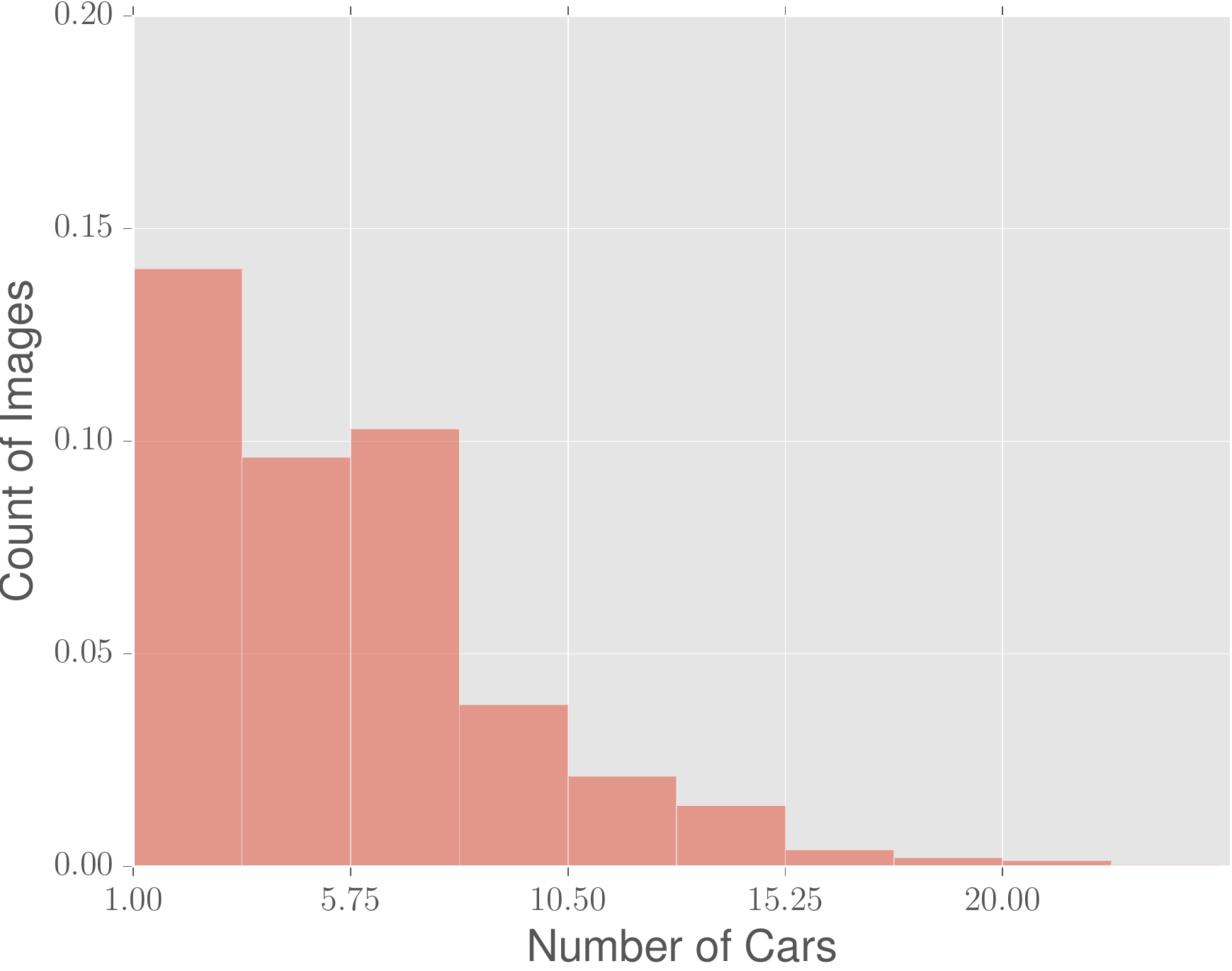}
\caption{Sim}
\end{subfigure}
\begin{subfigure}{0.3\linewidth}
\includegraphics[width=\linewidth]{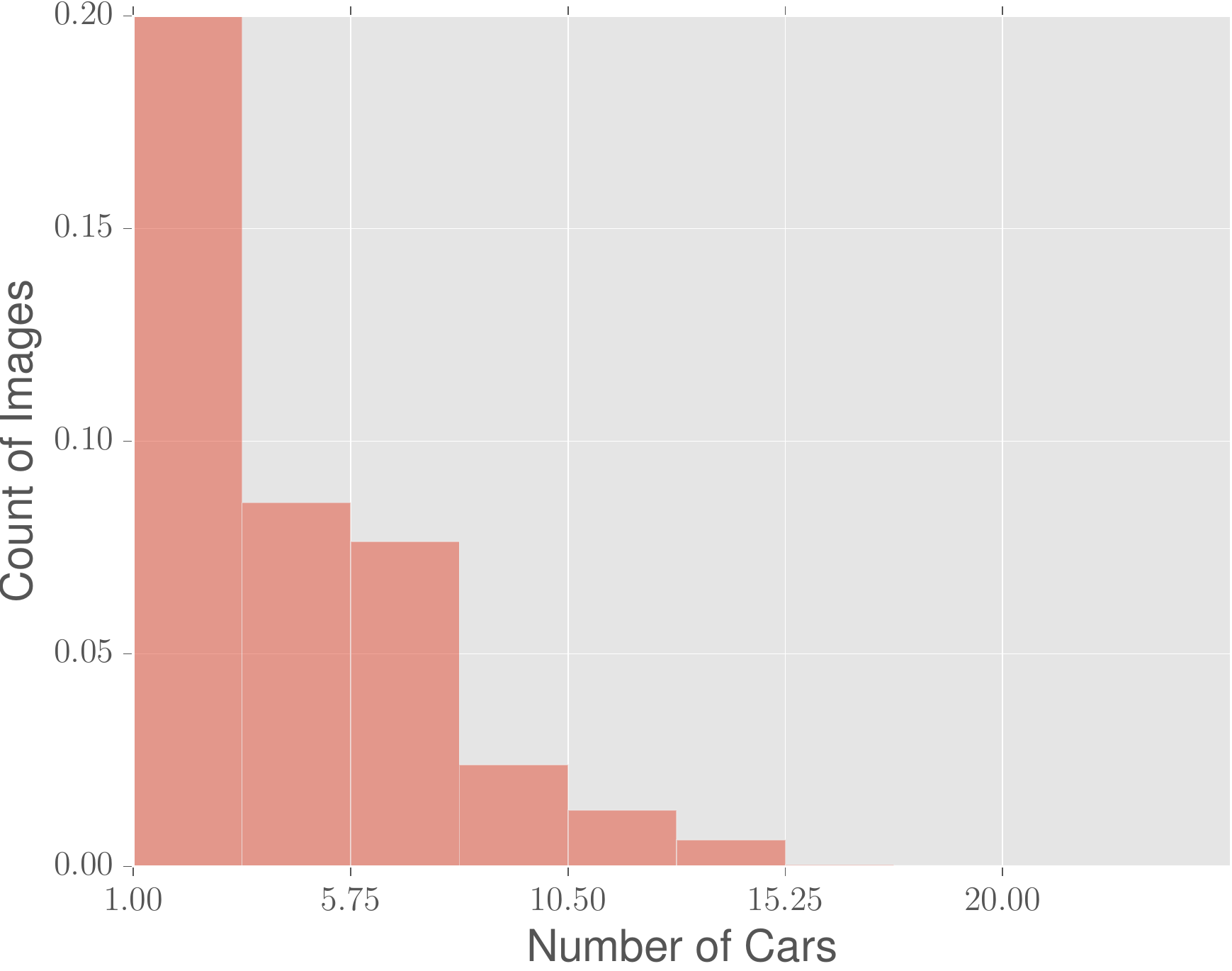}
\caption{KITTI}
\end{subfigure}
\caption{These figures depict histograms of the number of detections per frame in the Cityscapes, Simulation, and KITTI data sets. Note the similarity of the simulation and KITTI data set distributions. This may aid the network trained using the simulation data set when evaluated on the KITTI data set.}
\label{f:count}
\end{figure*}

To train the learning architecture described in the next subsection, we generated $3$ distinct simulated data sets of varying size: Sim $10$k with $10,000$ images, Sim $50$k with $50,000$ images, and Sim $200$k with $200,000$ images. The size of these data sets is compared to the state of the art human annotated data sets of real-world scenes in Table \ref{t:size_datasets}. A heatmap visualizing the distribution of the centroids of the car bounding boxes for the real-world scenes and the simulated scenes is illustrated in \figref{f:heatmap}. Notice the much larger spread of occurrence location in the simulated data than in the real world data set. Additionally \figref{f:count} depicts the distribution of the number of detections per frame. Notice that KITTI and the simulation data set are similarly distributed in terms of the number of detections per frame.

\begin{table}[htb]
    \centering
   \begin{tabular}{*2l} \toprule
\bfseries Data set & \bfseries \# of images \\ \midrule
Cityscapes~\cite{Cordts:2016aa} & 2,975 \\
KITTI~\cite{kitti} & 7,481 \\
Sim 10k & 10,000 \\
Sim 50k & 50,000 \\
Sim 200k & 200,000 \\
\end{tabular}
    \caption{This table presents the sizes of the data sets used for training and validation. Since one of the primary focuses of this paper is to understand data set bias in automated detection of objects for autonomous driving using vision, the performance of the network is evaluated on a data set that is distinct from the data set used to train the network. Note the relatively small sizes of the major real-world data sets used for training. }
    \label{t:size_datasets}
\end{table}

\subsection{Object Detection Network Architecture}\label{s:network}

Since a specific network architecture is not the focus of this paper, we employ a state of the art deep learning object detection network architecture: Faster-RCNN implemented in the Mx-Net framework~\cite{faster-rcnn,Chen:2015aa}. 
Full details of the approach appear in the paper, however the following important departures from the original reference implementation were made: 1)
use of VGG-16~\cite{vgg16} as opposed to AlexNet used in the original paper~\cite{faster-rcnn}.

The training and evaluation procedures are standard for an end-to-end training of Faster-RCNN. Layers copied from the VGG-16 network were initialized with Imagenet~\cite{imagenet} pre-trained weights in all cases. 

\section{Experimental Design}\label{s:exp}

All the network performance assessments are calculated by testing on real images from the KITTI data set. All $7481$ images in the KITTI ``training" set were used as the testing data for the proposed networks trained on any of the simulation or Cityscapes data sets.
Since each data set varies in size so dramatically, there was no fixed number of total iterations that all of the networks could be trained on to achieve optimal performance. Bearing this in mind, each network was run until performance asymptoted. Using this criteria, the network trained on Cityscapes was run for $13$ epochs, Sim 10k for $12$ epochs, Sim 50k for $12$ epochs and Sim 200k $8$ epochs. Note that an epoch represents one complete run through the entire data set. Training began with a learning rate of $10^{-3}$ and decreased by a factor of $10$ after every $10,000$ iterations until a minimum learning rate of $10^{-8}$ was achieved. We used a momentum factor of $0.9$ and a weight decay factor of $5\times10^{-4} $ along with a batch size of one image per \ac{GPU} for all experiments.

The performance of each of the trained networks was evaluated using a standard \ac{IoU} criteria for produced bounding boxes~\cite{kitti}. We use a $0.7$ \ac{IoU} overlap threshold as is typical in KITTI evaluation on cars. In training the network on the Cityscapes data, we used rectangular bounding boxes that were obtained by taking the smallest bounding box that fully enclosed the ``coarse'' pixel annotations provided in the data set.
Detections which are fully visible while having a minimum bounding box height of 40 pixels and a maximum truncation of 15\% are categorized as Easy. Detections which are partially occluded while having a minimum bounding box height of 25 pixels with a maximum truncation of 30\% are categorized as Moderate. Detections which are difficult to see with a maximum truncation of 50\% and a minimum bounding box height of 25 pixels are categorized to be Hard. 


\section{Results}\label{s:res}

\begin{figure}[htb]
    \centering
    \includegraphics[width=\linewidth]{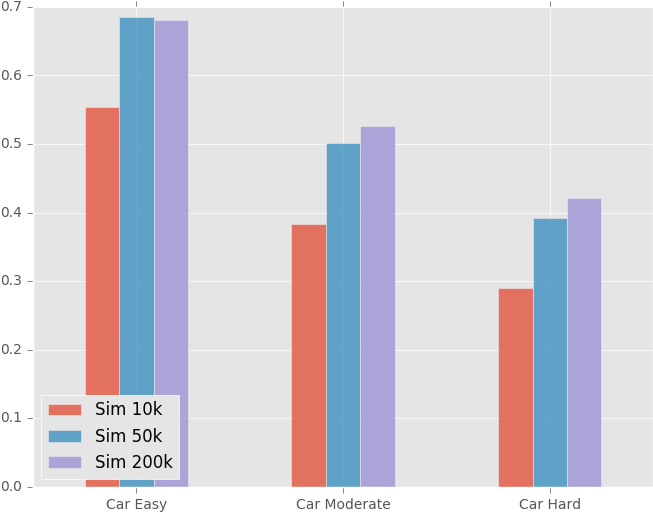}
    \caption{Mean average precision for testing on the KITTI data set with results reported on the Easy, Moderate and Hard divisions of the labels. Note we use all 7,481 KITTI training images for validation. We do this as no KITTI imagery is used for training. We are focused on understanding how generalizable networks trained on other data is to the general problem of object classification.}
    \label{fig:main_results}
\end{figure}

The goal of this work is to highlight the power of simulation and 3D engines as tools for the training of deep learning approaches for object classification. To this end \figref{fig:main_results} depicts the results of applying a typical Faster R-CNN (see \secref{s:network}) network trained on 10,000, 50,000 or 200,000 simulation images. 
First, it is important to notice that the base number of simulation images used is much larger than the real image data sets (KITTI and Cityscapes). It appears the variation and training value of a single simulation image is lower than that of a single real image. The lighting, color, and texture variation in the real world is greater than that of our simulation and as such many more simulation images are required to achieve reasonable performance. However, the generation of simulation images requires only computational resources and most importantly does not require human labeling effort. Once the cloud infrastructure is in place, an arbitrary volume of images can be generated. Second, there is a significant jump between 10,000 and 50,000 images. This indicates to us that there is a threshold of images above which the network learns a much more discriminative model of cars in the world. Perhaps most intriguing, we see continued improved performance between 50,000 and 200,000 images (albeit a much smaller jump). To confirm this is not simply a function of higher iteration count we ran 50,000 images through additional training epochs and actually saw a decrease in performance (most likely due to overfitting to that specific data). 

\begin{figure}[htb]
\begin{subfigure}{\linewidth}
\includegraphics[width=\linewidth]{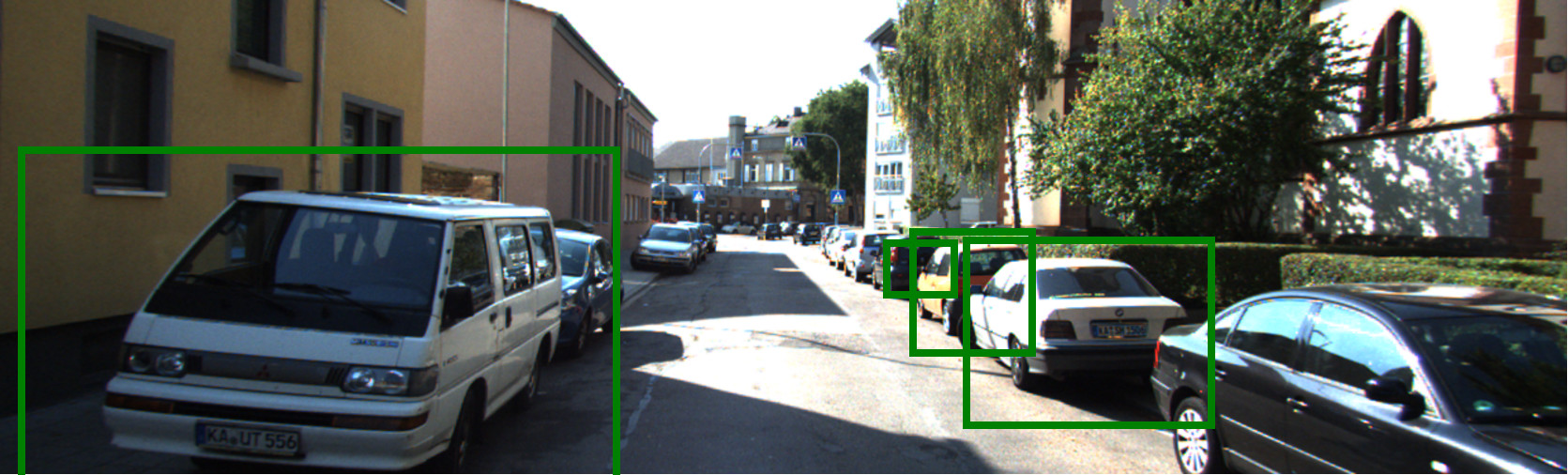}
\caption{10k}
\end{subfigure}
\begin{subfigure}{\linewidth}
\includegraphics[width=\linewidth]{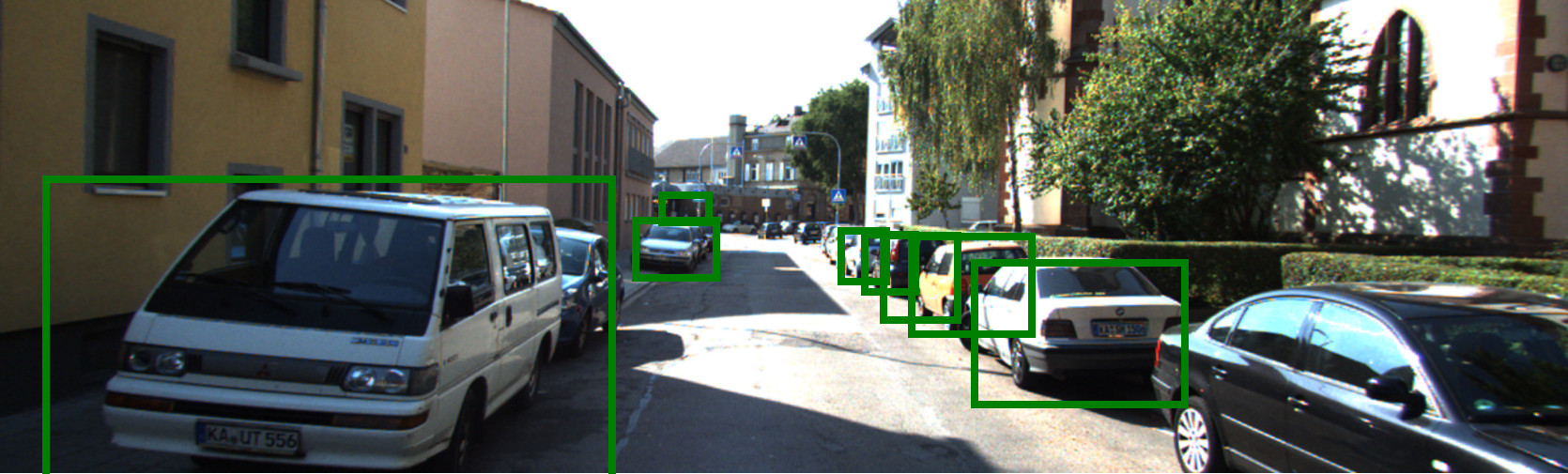}
\caption{50k}
\end{subfigure}
\begin{subfigure}{\linewidth}
\includegraphics[width=\linewidth]{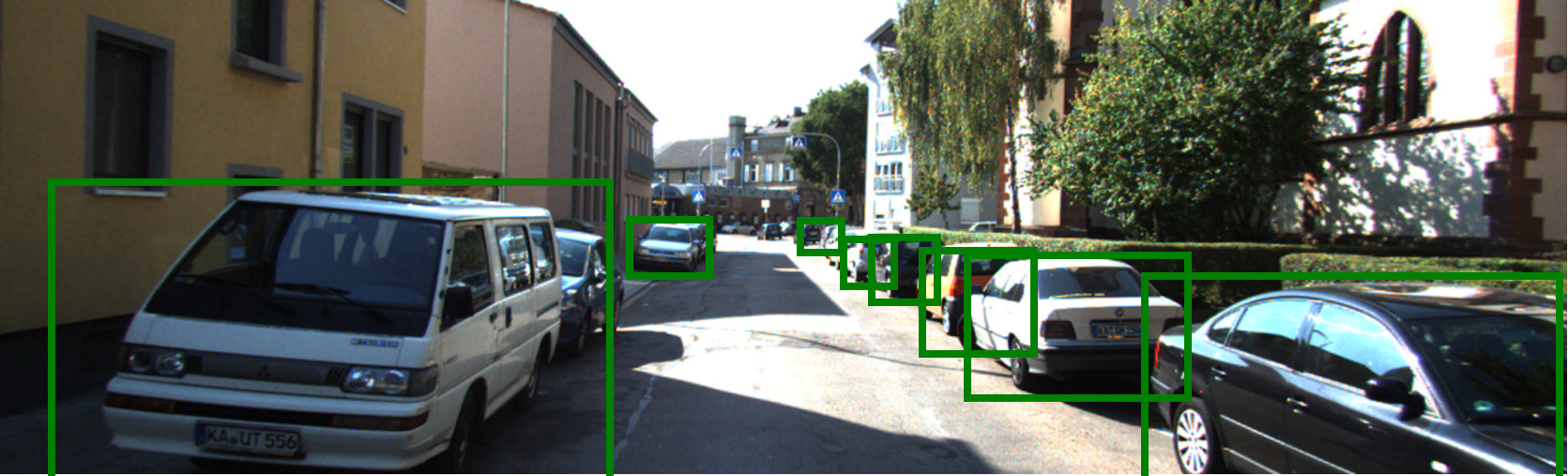}
\caption{200k}
\end{subfigure}
\caption{A visualization of some representative detections from training on three increasing volumes of simulation data. Note how the performance improves in the larger training  data sets. This is evaluated quantitatively in Table \protect\ref{t:datasets}, but the qualitative differences are illustrated in this figure.}
\label{f:qual-detections}
\end{figure}

In \figref{f:qual-detections}, we see the qualitative results of increasing the number of training images in the simulation data set. Many smaller arguably more challenging cars that appear at greater distance in the scene are detected by the simulation networks trained on more images. The continued improvement with additional simulation images points to an interesting discussion point: perhaps we are limited by data set size as opposed to network architecture. 

\begin{figure}[ht]
\begin{subfigure}{\linewidth}
\includegraphics[width=\linewidth]{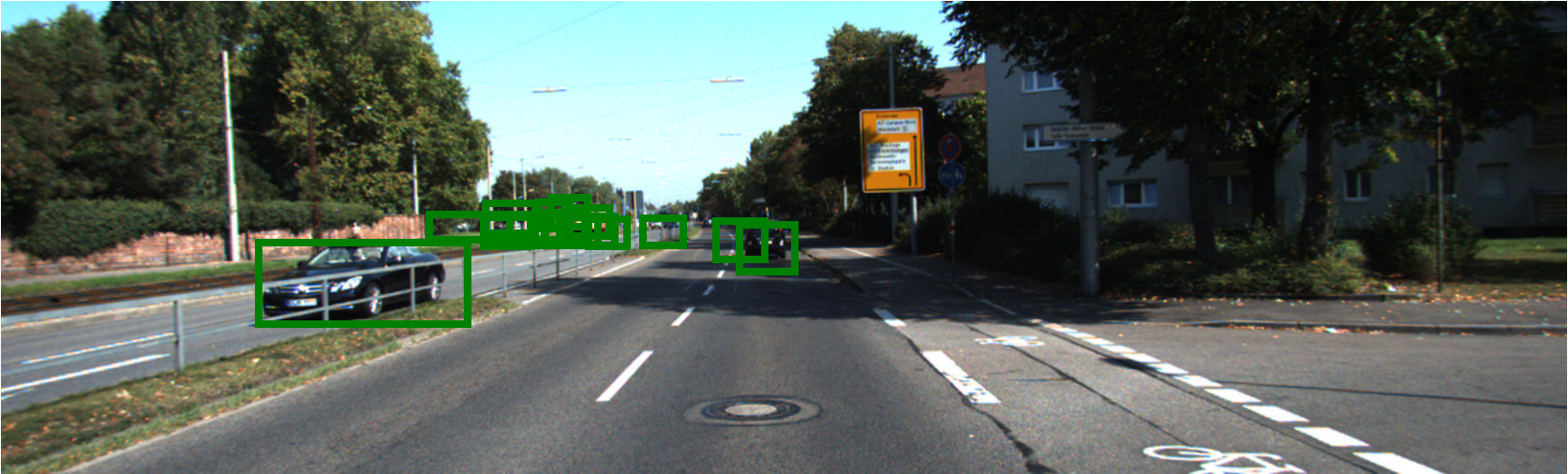}
\caption{Cityscapes}
\end{subfigure}
\begin{subfigure}{\linewidth}
\includegraphics[width=\linewidth]{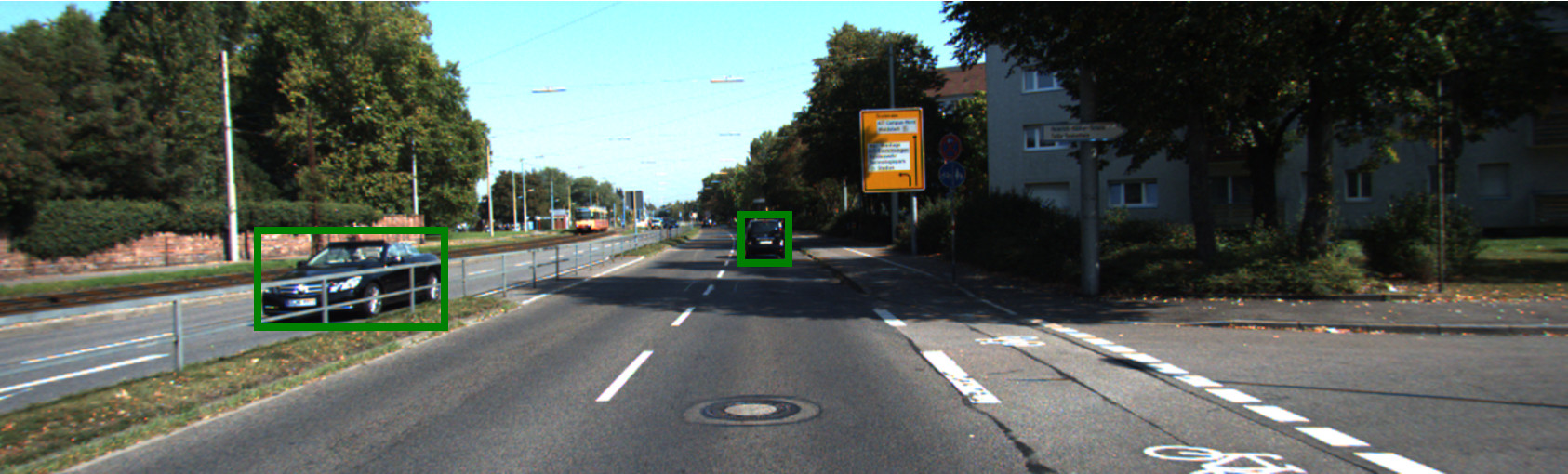}
\caption{200k}

\end{subfigure}

\caption{A visualization of some representative detections from training on Cityscapes and Sim 200k data sets. As seen in the figure, Sim 200k produces less cluttered outputs and better detections overall compared to Cityscapes.}
\label{f:qual-realsim}
\end{figure}

Beyond understanding the relative performance of varying numbers of simulation images it is important to understand how the simulation results relate to the performance on a network trained with annotated real images. Table~\ref{t:datasets} shows the results of both simulated trained networks, but also a network trained on real images drawn from a distinct data set (in this case Cityscapes). Notice that with larger number of only simulation images, we achieve superior performance to a network trained with only real images. In the table it can be seen that the Sim 50k and 200k datasets clearly outperform Cityscapes in the Easy, Moderate and Hard categories.  
Additionally, in \figref{f:qual-realsim} one can see the superior quality of the detection bounding boxes produced by the simulation trained network to those of the Cityscapes trained network.
In particular, the simulation trained model produces less cluttered outputs. Code and data for reproducing the results is available at \url{https://github.com/umautobots/driving-in-the-matrix} 


\begin{table}[htb]
    \centering
   \begin{tabular}{*4l} \toprule
\bfseries Data set &\bfseries Easy & \bfseries Moderate &  \bfseries Hard \\ \midrule
Sim 10k & 0.5542 & 0.3828 &0.2904 \\
Sim 50k & \textbf{0.6856} & 0.5008 & 0.3926 \\
Sim 200k & 0.6803& \textbf{0.5257}& \textbf{0.4207} \\
Cityscapes~\cite{Cordts:2016aa} & 0.6247 & 0.4274 & 0.3566 \\

\end{tabular}
    \caption{This table presents the results of the \ac{mAP} for $0.7$ \ac{IoU} on Easy, Moderate, and Hard Cars using a network never shown a KITTI image, but rather trained on Simulation and Cityscapes images respectively and then tested on all 7,481 KITTI images. Note our proposed approach, using only simulated car images, out performs real imagery (Cityscapes) on labels of all difficulties.}
    \label{t:datasets}
\end{table}

\section{Discussion}\label{s:disc}
The results show that a simulation only training approach is viable for the purpose of classifying real world imagery. 
This is a powerful concept for self-driving car research, particularly the prospect of using a simulation environment to improve training and testing for increasingly larger and more widely deployed fleets of vehicles. 
The idea of using training data from simulation for deep learning has been explored in prior work, but it's impact has been mitigated due to the mixing of simulated and real images. 
In addition, we believe the model of using traditional training/test splits for modestly sized data sets may be leading to undesirable local minima. 
More explicitly, this pattern of training may be leading to overfitting or even memorization, which may hinder the development of generalizable models for object recognition. 

The results achieved by the Cityscapes trained network, when evaluated on the KITTI data set, may in fact be due to overfitting since both data sets are from Germany and share similar cars and road structure. Though weather conditions varied marginally between the pair of data sets, they were both captured at the same time of day. Many open questions remain about the performance of deep learning networks when tested in much more diverse locations and weather conditions, as would be necessary for wide deployment.



\section{Conclusions \& Future Work}\label{s:con}
This paper presents a pipeline to gather data from a modern visual simulator with high realism to use as training data for object identification with deep learning networks. 
Networks trained using the simulated data were capable of achieving high levels of performance on real-world data without requiring any mixing with real-world training data set imagery. We highlight issues with data set bias in the way we train car detectors on modestly sized real data sets. 
Finally, we presented results that demonstrated improved performance from high numbers of simulated examples. 

Future avenues for research include: confirming the performance of the simulated data across many different deep learning network architectures, deepening network architectures to maximize the impact of larger numbers of training examples, and using active learning or other approaches to maintain the high levels of performance with smaller subsets of simulation imagery. The goal of smaller data sets will both reduce training time and distill training image importance and help us to understand redundancy in this type of training data. Finally, we would like to explore greater complexity and specificity in the generation of simulation images. 




\renewcommand{\bibfont}{\normalfont\small}
\printbibliography

\end{document}